\documentclass[letterpaper, 10 pt, conference]{IEEEconf}  % Comment this line out if you need a4paper

\IEEEoverridecommandlockouts                              % This command is only needed if 
\usepackage{graphicx} % for pdf, bitmapped graphics files
\usepackage[absolute,overlay]{textpos}
\usepackage{xcolor} % for \textcolor
\usepackage{float}
\usepackage{pifont} % for checkmark and x
\usepackage{amssymb}
\usepackage{amsmath}
\usepackage{multirow}
\usepackage{algorithm}
\usepackage{algpseudocode}

\usepackage{amsthm}

\newtheorem{theorem}{Theorem}
\newtheorem{lemma}{Lemma}      % Will be numbered independently
\newtheorem{property}{Property}

\let\labelindent\relax
\usepackage{enumitem}
\usepackage{array}
\usepackage{booktabs}

\title{\LARGE \bf
Uncertainty-Aware Multi-Robot Task Allocation \\ With Strongly Coupled Inter-Robot Rewards
}

\author{Ben Rossano$^{1,3}$, Jaein Lim$^2$, Jonathan P. How$^1$% <-this % stops a space
\thanks{$^{1}$B. Rossano and J.P. How are with the Aerospace Controls Lab, Massachusetts Institute of Technology, Cambridge, MA, USA {\tt\small \{brossano, jhow\}@mit.edu}}%
\thanks{$^{2}$J. Lim is with the Charles Stark Draper Laboratory, Cambridge, MA, USA {\tt\small {jlim}@draper.com}}%
\thanks{$^3$ B. Rossano is a Draper Scholar with the Charles Stark Draper Laboratory, Cambridge, MA. The authors would like to thank the Draper Scholars program for supporting this work.}
}

\usepackage[noadjust,sort,compress]{cite}

\makeatletter
\let\NAT@parse\undefined
\makeatother

\usepackage[colorlinks=true,linkcolor=black,citecolor=black,urlcolor=blue]{hyperref}
\makeatletter
\let\orglabel\label
\renewcommand{\label}[1]{\orglabel{#1}\hypertarget{#1}{}}
\makeatother

\begin{document}

\maketitle
\thispagestyle{empty}
\pagestyle{empty}
%%%%%%%%%%%%%%%%%%%%%%%%%%%%%%%%%%%%%%%%%%%%%%%%%%%%%%%%%%%%%%%%%%%%%%%%%%%%%%%%
\begin{abstract}
Allocating tasks to heterogeneous robot teams in environments with uncertain task requirements is a fundamentally challenging problem. Redundantly assigning multiple robots to such tasks is overly conservative, while purely reactive strategies risk costly delays in task completion when the uncertain capabilities become necessary. This paper introduces an auction-based task allocation algorithm that explicitly models task requirement uncertainty, leveraging a novel \textit{strongly coupled} formulation to allocate tasks such that robots with potentially required capabilities are naturally positioned near uncertain tasks. This approach enables robots to remain productive on nearby tasks while simultaneously mitigating large delays in completion time when their capabilities are required. Through a set of simulated disaster relief missions with task deadline constraints, we demonstrate that the proposed approach yields up to a 15\% increase in expected mission value compared to redundancy-based methods. Furthermore, we propose a novel framework to approximate uncertainty arising from \textit{unmodeled} changes in task requirements by leveraging the natural delay between encountering unexpected environmental conditions and confirming whether additional capabilities are required to complete a task. We show that our approach achieves up to an 18\% increase in expected mission value using this framework compared to reactive methods that do not leverage this delay.

\end{abstract}
\vspace{-0.05cm}
%%%%%%%%%%%%%%%%%%%%%%%%%%%%%%%%%%%%%%%%%%%%%%%%%%%%%%%%%%%%%%%%%%%%%%%%%%%%%%%%
\section{Introduction} \label{sec:introduction}
Heterogeneous multi-robot systems can achieve higher levels of efficiency and effectiveness than individual robots by leveraging task parallelization and diverse sets of capabilities \cite{doi:10.1177/0278364904045564}. Such systems are particularly well-suited for large-scale missions, such as search and rescue \cite{7084641}, disaster relief \cite{jones_time-extended_2011}, and environmental monitoring \cite{cao2013multi}, that require operating in partially known environments. The success of these missions relies heavily upon multi-robot task allocation (MRTA), the process of determining how to assign tasks to robots to best exploit the team's capabilities. 

\begin{figure}[t]
    \centering
    \includegraphics[width=0.45\textwidth]{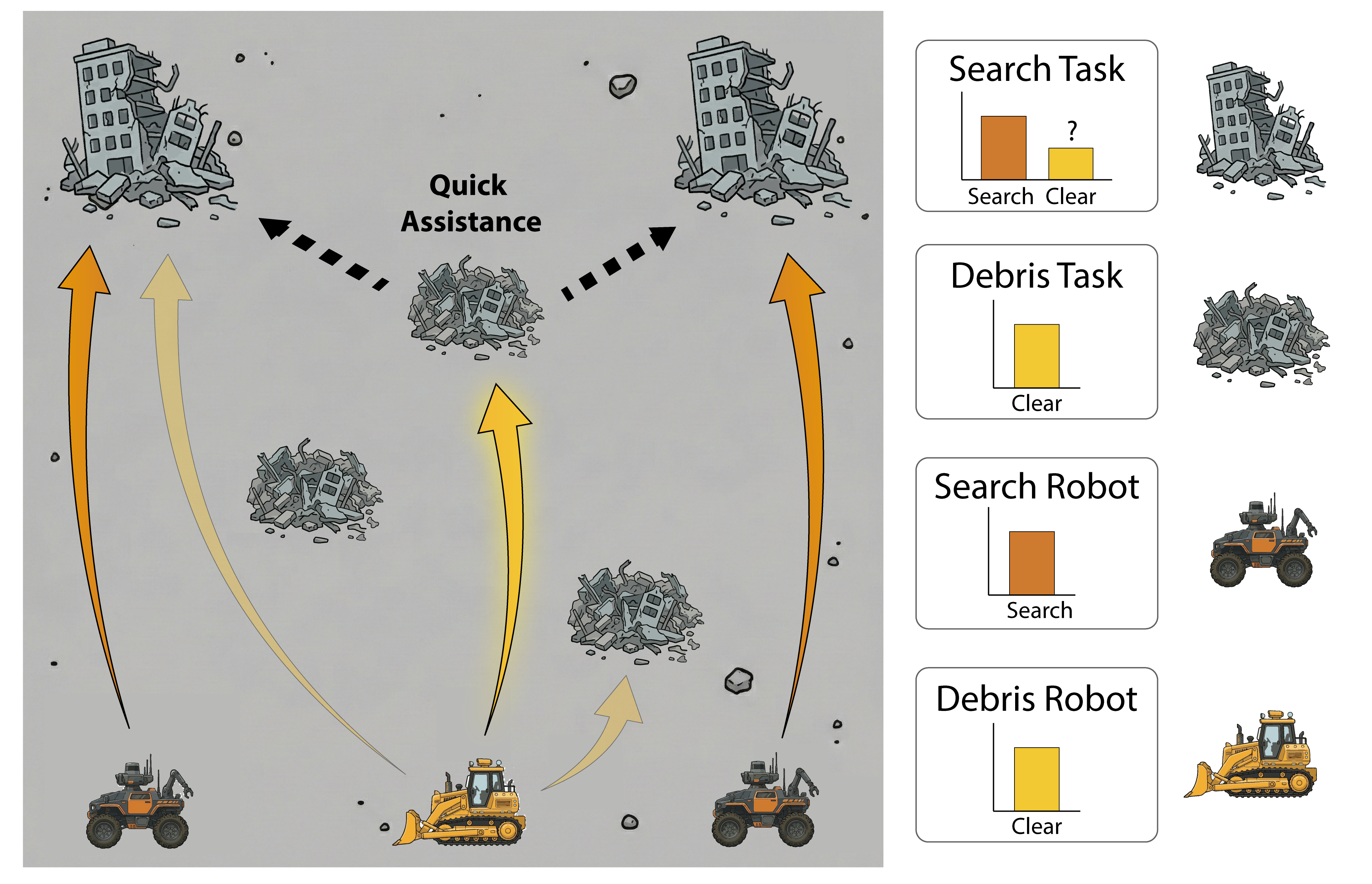}
    \vspace{-0.3cm}
    \caption{A disaster relief scenario with task requirement uncertainty. Debris tasks can always be completed by a debris robot, whereas search tasks require a search robot and may require additional support from a debris robot with some probability. Our approach strategically selects a more distant debris task, balancing both timely debris task completion and quick assistance for either search task.}
    \label{fig:disaster_example}
    \vspace{-0.6cm}
\end{figure}

When environments are only partially known, it is often challenging to determine which robot capabilities will ultimately be required for successful task completion. For instance, consider the disaster relief mission in Fig. \ref{fig:disaster_example} consisting of a set of high-value search tasks and comparatively lower-value debris-cleanup tasks. While search tasks primarily require the ``search" capability, debris may be present that prevents task completion without a larger debris-clearing robot first removing the obstruction. In such scenarios, MRTA becomes a strategic challenge: allocation must optimize the positioning of heterogeneous robots to maximize performance while hedging against the risk of unfulfilled requirements.

Existing approaches that address uncertain or changing task requirements typically fall into two broad categories. When uncertainty is known a priori, teams must exhibit \textit{robustness} by proactively modifying pre-execution plans to anticipate possible outcomes \cite{robust_dynamic_allocation}. This is generally achieved through redundancy \cite{9381653}, which produces overly conservative solutions, or risk avoidance \cite{Ponda2012Thesis}, which cannot proactively hedge against uncertainty when deferring uncertain tasks is undesirable. When task requirements may change unexpectedly, robots must exhibit \textit{resilience}, adapting in real time to overcome unanticipated conditions \cite{resilient-flocking}. Existing methods achieve this through failure detection and task progress monitoring \cite{ALLIANCE, mayya_resilient_2021}, often first allowing robots to attempt a local recovery before task requirements are formally updated (e.g., a search robot encountering debris may first explore alternative routes to resolve the situation itself before a dedicated debris robot is requested). This opens a window of time between the team's initial detection of uncertainty and any subsequent task requirement updates, which can be leveraged to proactively replan.

This paper proposes a novel unified framework for MRTA that supports both robust and resilient mission execution. We first introduce a method to approximate task requirement uncertainty that arises from unexpected conditions. Specifically, when a robot is attempting a local recovery, there is inherent uncertainty about the capabilities required to complete the task---referred to as unmodeled task requirement uncertainty. To estimate this uncertainty, we leverage the discrepancy between expected and observed task progress as a signal that additional capabilities may be required.

Leveraging the uncertainty models available in both settings, we present AURA (Auction-based Uncertainty-aware Robot Allocation), an algorithm that explicitly captures inter-robot dependencies caused by task requirement uncertainty. Our key idea is to \textit{adjust task utilities to reflect not only the intrinsic value of a task but also the potential benefit of providing assistance for uncertain tasks}. This design incentivizes robots with potentially required capabilities---referred to as support robots---to remain productive on nearby tasks while simultaneously positioning themselves to assist uncertain tasks if needed. The resulting MRTA formulation is particularly challenging due to two inter-robot dependencies that arise: (i) the utility of some tasks depends on the location and expected execution time of nearby uncertain tasks, and (ii) while all support robots benefit from positioning near uncertain tasks, only one will ultimately be reassigned if assistance is needed. Efficient allocation therefore requires explicitly modeling this coupling to avoid over-incentivizing robots to cluster around the same uncertain tasks.
Our work can be summarized by three primary contributions:
\begin{itemize}
    \item We propose a unified framework for task requirement uncertainty that supports both modeled uncertainty and a novel approximation of unmodeled uncertainty. 
    \item We present an auction-based, uncertainty-aware algorithm, AURA, that strategically assigns tasks to position support robots near uncertain tasks. AURA explicitly captures inter-robot dependencies that existing methods fail to model while guaranteeing convergence and pseudo-polynomial time complexity.
    \item We demonstrate that under modeled uncertainty, AURA reduces the expected performance loss due to missed deadlines by up to 15\% compared to redundancy-based approaches. Furthermore, we show that approximating unmodeled uncertainty for proactive replanning improves performance by up to 18\% compared to reactive replanning after task requirements are formally updated.

\end{itemize}

\section{Background and Related Work} \label{sec:rel_work}

\subsection{Auction-Based Approaches} 
Auctions have been widely adopted in MRTA due to their efficiency, scalability, and natural suitability for decentralized systems \cite{quinton_market_2023}. Among these, the Sequential Single-Item (SSI) auction \cite{koenig_power_nodate} and the Consensus-Based Bundle Algorithm (CBBA) \cite{choi_consensus-based_2009} are fundamental methods. Notably, auctions provide a 50\% optimality guarantee under the condition of submodularity, which requires that the value of adding a task to the allocation satisfies a diminishing marginal gain \cite{nemhauser1978analysis}. 

Many realistic MRTA problems, however, violate this submodularity assumption. Our approach falls into a broader category of problems in which task utilities depend on assignments across multiple robots, known as cross-schedule dependencies (XDs) \cite{doi:10.1177/0278364913496484}. In these settings, the value of a task may increase once other tasks are added to the team allocation, removing performance guarantees and compromising convergence in decentralized frameworks that rely on local consensus \cite{choi_consensus-based_2009}. In \cite{SSITemporal} and \cite{whitten_decentralized_2011}, SSI and CBBA are extended, respectively, to support temporal and precedence constraints between tasks. However, these methods rely on a set of known constraints. In this work, constraints may only arise during execution (i.e., when task requirements change), making it impossible to specify them a priori.

\subsection{Robust Task Allocation}

Robust MRTA methods address uncertainty that is known a priori, enabling proactive allocation decisions that anticipate variability in robot capabilities or task requirements \cite{chaari_scheduling_2014}. Prorok \cite{prorok-redundant} redundantly assigns multiple robots to the same task under travel time uncertainty to maximize the probability of timely task completion. Fu et al. \cite{fu_robust_2022} consider uncertainty in both robot capabilities and task requirements, formulating a stochastic program that minimizes mission risk. However, their approach also relies on over-assigning multiple robots to mitigate risk. Ponda \cite{Ponda2012Thesis} extends CBBA with a sampling-based variant that incorporates probabilistic mission parameters. However, it lacks the ability to proactively hedge against uncertain cross-schedule dependencies. 

\subsection{Resilient Task Allocation}

Resilient MRTA methods address unexpected scenarios such as robot failures and unforeseen changes in task feasibility \cite{prorok_beyond_2021}. These approaches typically rely on detecting disruptions and reallocating tasks in response. Neville et al. \cite{neville_d-itags_2022} address resilience through targeted repairs of existing solutions rather than full-scale reallocations. However, their method assumes that failures occur without the opportunity for local recovery. Notomista et al. \cite{notomista_resilient_2021} use deviations between expected and observed task progress to detect external disturbances. However, reallocation is triggered only after a deviation threshold is met, missing the opportunity for proactive reallocation during earlier stages of task degradation.

\allowdisplaybreaks

\section{MRTA Under Uncertainty Formulation}
Consider an environment containing a set of $N_r$ robots and $N_t$ tasks. We denote the set of robots as $\mathcal{I} \triangleq \{1,\ldots,N_r\}$ and the set of tasks as $\mathcal{J} \triangleq \{1,\ldots,N_t\}$. 
The goal of the multi-robot task allocation problem with uncertainty is to find a conflict-free assignment of robots to tasks that maximizes some stochastic objective function. Using expected value as our stochastic metric, the task allocation problem can be written as the following integer program:

\begin{equation}
\begin{aligned}
\max_{\mathbf{x}, \mathbf{p}} \quad & 
\mathbb{E}_{\theta} \left[ 
\sum_{i=1}^{N_r} \left(\sum_{j=1}^{N_t} 
c_{ij}\!\left(\mathbf{x}, \mathbf{p}, \theta \right) \mathbf{x}_{ij} \right)\right] \\
\text{s.t.} \quad 
& \sum_{j=1}^{N_t} \mathbf{x}_{ij} \leq L_t, \quad \ \forall i \in \mathcal{I}, \\
& \sum_{i=1}^{N_r} \mathbf{x}_{ij} \leq 1, \quad \ \ \ \forall j \in \mathcal{J}, \\
& \mathbf{x}_{ij} \in \{0,1\}, \quad \ \ \ \forall (i,j) \in \mathcal{I} \times \mathcal{J}, \\
\end{aligned}
\end{equation}
where $\mathbf{x}$ is a binary allocation matrix ($\mathbf{x}_{ij} =1$ if robot $i$ is assigned to task $j$), $L_t$ is the maximum number of tasks each robot can be assigned, $\mathbf{p}$ is a matrix representing all robot paths, $\theta$ is a vector of uncertainty parameters, and $c_{ij}$ is the reward function for robot $i$ completing task $j$. Note that $c_{ij}$ depends on all assignments and paths, not just those of robot $i$, when modeling inter-robot dependencies.

\subsection{Uncertainty Model}
We model task requirement uncertainty as a Bernoulli random variable, yielding a simple binary representation. This can be parameterized by $\theta \triangleq \{(\mathcal{P}_1, \mathcal{T}_1),\ldots,(\mathcal{P}_{N_t}, \mathcal{T}_{N_t})\}$, where $\mathcal{P}_j$ is the probability that task $j$ will require an additional capability (e.g., debris-clearing) and $\mathcal{T}_{j}$ is the amount of time after arrival that this is expected to be discovered. While we focus on scenarios in which tasks require at most one additional capability, this framework can be extended to support multiple uncertain requirements. 

In the robust framework, we assume that $\theta$ is known prior to mission execution. In the resilient framework, task requirement uncertainty is determined by measuring changes in the expected and observed task progress over time. We reinterpret the robot-task specialization metric introduced in \cite{notomista_resilient_2021} as a measure of task feasibility, $f(t):\mathbb{R}_{\geq0}\rightarrow[0,1]$, to quantify uncertainty once the robot detects that an additional capability \textit{may} be required for successful task completion. We denote the expected and observed task progress at discrete time $t_k$ by $\phi_{\text{exp}_j}(t_k)$ and $\phi_{\text{obs}_j}(t_k)$, respectively. For a task $j$ being executed, the expected progress is defined as:
\begin{equation}
    \phi_{\text{exp}_j}(t_k) = \phi_{\text{obs}_j}(t_{k-1})+\phi_{\text{step}_j},
\end{equation}
where $\phi_{\text{step}_j}$ is the expected change in progress during a discrete time step. This enables us to compare the expected progress with the observed progress at time $t_k$:
\begin{equation}
    \Delta\phi(t_k)= \phi_{\text{exp}_j}(t_k) - \phi_{\text{obs}_j}(t_k),
\end{equation}
assuming that robots are capable of identifying their current true progress (see \cite{notomista_resilient_2021} for examples). We can then model how task feasibility decays over time, which ultimately informs our uncertainty estimate. Starting from an initial feasibility value of $f_j(0)=1$, our uncertainty model is defined incrementally as:
\begin{align}
    f_{j}(t_{k+1}) &= f_{j}(t_k)-\beta\Delta\phi(t_k), \\
    \mathcal{P}_j(t) &=  1 - \gamma f_{j}(t_{k+1}), \\
    \mathcal{T}_j(t) &= f_{j}(t_{k+1})/\beta,
\end{align}
where $\beta$ is a decay parameter that dictates how quickly feasibility decreases, and $\gamma$ determines how strongly feasibility influences the approximated uncertainty. Note that we bound $f_j$ to ensure that it does not go below zero. Intuitively, task uncertainty can be interpreted as the inverse of feasibility, allowing us to capture resilience within an otherwise robust formulation. The remaining feasibility defines a local recovery window, with $\mathcal{T}_j$ denoting the time until $f_j=0$. During this period, the assigned robot is given the opportunity to recover before modifying the task requirements. Successful recovery resets $f_j=1$, eliminating the uncertainty, whereas reaching $f_j=0$ marks the end of the recovery window, triggering a formal update of the task requirements.

\subsection{Auxiliary Tasks} \label{subsec:aux}
In scenarios with high-uncertainty tasks and no nearby tasks to position a support robot, redundantly assigning support robots to uncertain tasks may be an effective strategy to maximize the expected mission reward. Therefore, we introduce auxiliary tasks requiring the support capability at the locations of all uncertain tasks \cite{whitten_decentralized_2011}. Auxiliary tasks have no intrinsic value, but robots assigned to them receive rewards for expected future support, similar to the actual tasks.

\section{Proposed Auction-Based Approach}
In this section, we propose a centralized auction-based algorithm that extends SSI \cite{koenig_power_nodate} to support coupled inter-robot rewards. In particular, we conduct auctions where only the task with the single highest bid is added to the allocation, though tasks are not permanently removed from the available task list. This enables rebidding, which is crucial given that bid values change dynamically as other robots’ assignments evolve. Inspired by \cite{choi_consensus-based_2009}, we store two types of task lists for each robot: the bundle $\mathbf{b_i}$ and the path $\mathbf{p}_i$. Tasks in a bundle are listed in the order that they were added to the bundle, while tasks in a path are listed in the order that they will be executed. 

Consistent with the motivating example in Section \ref{sec:introduction}, we establish that tasks with requirement uncertainty carry a larger intrinsic value. As such, we assume that a support request from any high-value, uncertain task (HVUT)---defined as the subset $\mathcal{H}\subset\mathcal{J}$---triggers the immediate reassignment of the closest available support robot. To explicitly model these potential future reassignments, we introduce two new lists for each robot: (i) $\mathbf{t}_i$, where $\mathbf{t}_{ik}$ is robot $i$’s projected arrival time at task $k\in\mathcal{H}$, and (ii) $\mathbf{r}_i$, where $\mathbf{r}_{ik} \in \{0,1\}$ denotes whether robot $i$ is assigned to the support of task $k$.

\subsection{Rewards and Scoring}
Let $R_i(\mathbf{p}, \mathbf{t}_i, \mathbf{r}_i)$ define the total reward robot $i$ receives for completing the tasks in $\mathbf{p}_i$ plus any additional reward for supporting HVUTs. This reward is defined for a given realization of $\theta$, where $\mathbf{r}_{i}$ and $\mathbf{t}_{i}$ are determined from a sample of task uncertainty outcomes (see Algorithm \ref{alg:essential_assignment}):

\begin{equation}
\begin{aligned}
R_i(\mathbf{p}, \mathbf{t}_i, \mathbf{r}_i) 
&= 
    \sum_{j \in \mathbf{p}_i} 
         \lambda^{\tau_i^j(\mathbf{p}_i)} \, \bar{v}_j \ \\
&\quad + \sum_{k \in \mathcal{H}} 
    \mathbb{I}(\mathbf{r}_{ik} = 1) \,
    \lambda^{\mathbf{t}_{ik}} \, \bar{v}_k \\[0.3em]
 \text{s.t.} \hspace{0.5cm} & \quad \  \tau_i^j \leq \mathcal{D}_j, \quad \forall j \in \mathcal{J}.
\end{aligned}
\end{equation}
The first term captures the reward for executing each task in $\mathbf{p}_i$, where $0 < \lambda < 1$ is a time-discount factor, $\tau_i^j(\mathbf{p}_i)$ is robot $i$'s completion time of task $j$ given its path, and $\bar{v}$ is a task's intrinsic value. The second term accounts for support rewards, which are only received if robot $i$ is chosen to support task $k$ at expected support time $\mathbf{t}_{ik}$. Lastly, $\mathcal{D}_j$ is the deadline of task $j$, after which no reward is given. 

Capturing the impact of inter-robot dependencies requires understanding how the insertion of a task into one robot’s path affects the reward of the entire team. To this end, we define a global score function $S$ that models the \textit{joint} reward of all robots:
\begin{equation}
S(\mathbf{p}, \mathbf{t}, \mathbf{r}) = \sum_{i \in \mathcal{I}}  R_i(\mathbf{p},\mathbf{t}_i, \mathbf{r}_i).
\end{equation}
This enables task bids to be evaluated as the joint marginal gain of the team for a new task assignment, greedily steering the system in the direction of increasing global reward.

\subsection{Iterative Single-Item Auction}
Algorithm \ref{alg:seq-bundle} presents the high-level pseudo-code for our approach. First, the best possible bid is calculated for each robot $i$, where $s_i^*$ corresponds to the bid value (measured as a marginal gain) and $j_i^*$ is the associated task. Then, the largest bid across the team is determined and added to the existing allocation if it exceeds a minimum threshold $\epsilon$. Enforcing a single-item auction ensures that accepted bids reflect their true marginal contribution to the global reward. Although multiple bids may \textit{individually} improve the current allocation, coupling between tasks can cause their combined marginal gain to be overestimated. This can result in a net negative contribution that decreases the global reward and violates the monotonicity assumption required for convergence (see Section \ref{sec: theory}).

% , preventing the overestimation that arises when multiple independently evaluated bids are accepted simultaneously in the presence of inter-robot dependencies. Such overestimation causes non-monotone global reward changes, violating a core assumption needed for convergence (see Section \ref{sec: theory}).

To encourage rebidding, bundles are constructed iteratively, adding tasks one position at a time and advancing only once the highest available bid drops below $\epsilon$. Since the marginal gain of inserting an unassigned task often outweighs the marginal gain of rebidding on an already assigned task, constructing bundles all at once can cause highly beneficial changes in task ownership to be overlooked. This iterative structure bounds the rate at which unassigned tasks can be added (at most $N_r$ per position), creating opportunities for rebidding through a task swapping mechanism detailed in the following subsection. We choose $\epsilon$ to be significantly smaller than the typical marginal gain of inserting an unassigned task to ensure that it does not induce premature bundle position advancement that would degrade solution quality. Rather, it serves to filter out negligible task swaps while preserving meaningful improvements.

\subsection{Task Swapping}
We introduce a task swapping mechanism, inspired by local search, that allows robots to bid on previously assigned tasks by swapping with their current task at the ongoing bundle position. If no such task exists (e.g., at the start of a new bundle position), the robot losing its task simply removes that task from its path.  The following example demonstrates task swapping when Robot 1 (first row) is bidding on a previously assigned task, Task 3, in Robot 2's path (second row):
\[
\mathbf{p}^{\text{prev}} = \begin{bmatrix}
2 & 4 & 6 \\
1 & 3 & 5
\end{bmatrix}
\xrightarrow{\text{swap}}
\mathbf{p} = \begin{bmatrix}
2 & 4 & 3 \\
1 & 6 & 5
\end{bmatrix}.
\]

Since bids are evaluated at the global level, we treat swaps the same as any other bid. 

\setlength{\textfloatsep}{10pt plus 1pt minus 2pt}
\begin{algorithm}[t]
\caption{Iterative Single-Item Task Assignment}
\label{alg:seq-bundle}
\begin{algorithmic}[1]
\small
\Procedure{SolveAssignment}{$\theta$}
    \State $\mathbf{b}, \mathbf{p}, \mathbf{t} \gets \emptyset, \emptyset, \emptyset$
    \State $k \gets 1$ \Comment{current bundle position}
    \While{$k < L_t$}
        \For{$i \in \mathcal{I}$}
            \State $c_{j} \gets \max_{n \le k} \textsc{GetBid}(\mathbf{p}, \mathbf{t},j, n, \theta), \hspace{0.07cm} \forall j \in \mathcal{J} \setminus \mathbf{b}_i$
            \State $j_i^* \gets \arg\max_j c_{j}$
            \State $s_i^* \gets c_{j_i^*}$
            % \State Compute $c_{ij^*}, n^*$ using Algorithm~\ref{alg:sampledscore}
            % \State Update $y_{ij^*}, z_{ij^*}$ if $c_{ij^*} > y_{ij^*}$
        \EndFor
        \State $i_{\text{best}} \gets \arg\max_i s_{i}^*$
        \If{$s_{i_{\text{best}}}^* < \epsilon$} \label{line:eps_marg_gain}
            \State $k \gets k + 1$
        \Else
            \State Update $\mathbf{b}, \mathbf{p}, \mathbf{t}$
            % \If{$\mathbf{p}^{\text{prev}} = \mathbf{p}$}
            %     \State $k \gets k + 1$
            % \EndIf
        \EndIf
    \EndWhile
    \State \Return $\mathbf{p}$
\EndProcedure
\end{algorithmic}
\end{algorithm}

\begin{algorithm}[t]
\caption{Bid Calculation Using Joint Marginal Gain}
\label{alg:sampling}
\begin{algorithmic}[1]
\small
\Procedure{GetBid}{$\mathbf{p}, \mathbf{t},j,n, \theta$}
    \State $\mathbf{p'} \gets \{\mathbf{p}_0,\ldots,\mathbf{p}_i \oplus_n \{j\},\ldots,\mathbf{p}_{N_r}\}$ 
    \State Swap tasks if necessary
    \State Update $\mathbf{t}_{ik}$ based on $\mathbf{p}'_i$ and $\mathcal{T}_k, \quad \forall k \in \mathcal{H}$
    \State $\{\theta_1,\ldots,\theta_{N_s}\} \sim f(\theta)$
    \State $\{w_1,\ldots,w_{N_s}\} \sim f(\theta)$
    \For{$s \in \{1,\ldots, N_s\}$}
        \State $\mathbf{r}, \mathbf{t} \gets \textsc{GetSupportInfo}(\mathbf{p}, \mathbf{t}, \theta_s)$ 
        \State $c_{j}^{(s)} \gets S(\mathbf{p'}, \mathbf{t}, \mathbf{r})-S(\mathbf{p}, \mathbf{t}^{\text{prev}}, \mathbf{r}^{\text{prev}})$ 
        \vspace{2pt}
    \EndFor
    \State $c_{j} \gets \sum_{s=1}^{N_{s}} w_s \, c_{j}^{(s)}$
    \State \Return $c_{j}$
\EndProcedure
\end{algorithmic}
\end{algorithm}

\subsection{Computing Bids}

Algorithm \ref{alg:sampling} describes the procedure for calculating the bid associated with inserting task $j$ into the $n^{\text{th}}$ position of robot $i$’s path. After updating the projected arrival times of robot $i$ for each HVUT, $N_s$ samples are generated from the joint task uncertainty distribution by combining individual task samples (assuming independence). Depending on the desired $N_s$, all possible outcomes may also be enumerated (i.e., when $2^{|\mathcal{H}|} < N_s$).

Before calculating the joint marginal gain of task $j$, $\mathbf{r}$ and $\mathbf{t}$ must be updated to reflect how each robot would respond to a given sample $\theta_s$ (Algorithm \ref{alg:essential_assignment}), where $N_h=|\mathcal{H}|$. This \textit{rollout} procedure enables the team to evaluate the true sequence of outcomes within that sample. First, HVUTs are filtered to include only those that require the additional support capability in this current sample, which we call failed tasks. These tasks are then sorted by their uncertainty realization times, which are a function of the primary robot's arrival time and $\mathcal{T}_k$. In order of realization time, support robots are greedily assigned. Lastly, $\mathbf{t}$ is updated to reflect the new projected arrival time for all future failing HVUTs. This allows accurate modeling of scenarios where the best behavior involves one robot supporting multiple HVUTs.

\begin{algorithm}[t]
\caption{Sequential Support Assignment}
\label{alg:essential_assignment}
\begin{algorithmic}[1]
\Procedure{GetSupportInfo}{$\mathbf{p}, \mathbf{t}, \theta_s$}
    \State $\mathbf{r}_{ik} \gets 0, \quad \forall i \in \mathcal{I}, \, k \in \mathcal{H}$ 
    \State $\mathcal{H}_{failed} \gets \{ k \mid \theta_k = 0 \}$ \Comment{Identify all failed tasks}
    \State $\mathcal{H} \gets \textsc{SortByRealizationTime}(\mathcal{H}_{failed}, \mathbf{p}, \theta_s)$ 
    \For{$u \gets 1$ to $N_h$}
        \State $k \gets \mathcal{H}_u$
        \State $i^* \gets \arg\min_i t_{ik}$ 
        \State $\mathbf{r}_{i^*k} \gets 1$ 

        \For{$n \gets u + 1$ to $N_h$}
            \State $k_{next} \gets \mathcal{H}_n$
            \State $\mathbf{t}_{i^*k_{next}} \gets \textsc{UpdateArrival}(i^*, k, k_{next}, \mathbf{t})$
        \EndFor
    \EndFor
    \State \Return $\mathbf{r}, \mathbf{t}$
\EndProcedure
\end{algorithmic}
\end{algorithm}

\section{Theoretical Analysis}
\label{sec: theory}
\subsection{Preliminaries}

\begin{property}
\label{prop: mono}
The single-item auction mechanism enforces a minimum improvement threshold $\epsilon > 0$. Formally, a new allocation $\mathcal{A}_{k+1}$ is accepted over the current allocation $\mathcal{A}_k$ if and only if the marginal gain satisfies:
\begin{equation}
    S(\mathcal{A}_{k+1}) - S(\mathcal{A}_k) \ge \epsilon.
\end{equation}

\end{property}

\begin{property}
\label{prop: syn}
We define the support reward for any task $j$ as \textit{exclusive}, meaning it is determined by a single robot with the earliest projected arrival rather than a sum of all available support robots. Formally, for an uncertain task $k$ and a set of support robots $\mathcal{I}_k$:
\begin{equation}
    R_{support}(k) = \max_{i \in \mathcal{I}_k} \left( \lambda^{\mathbf{t}_{ik}} \, \bar{v}_k \right).
\end{equation}
This implies that the support reward for any single task is bounded by a system constant $Q_{\max}$, regardless of the total number of robots $N_r$.
\end{property}

\subsection{Convergence Analysis}
\begin{lemma}
\label{lemma:boundedness}
The global objective function $S(\mathcal{A})$ is bounded from above by a finite value.
\end{lemma}

\begin{proof}
Since the set of tasks $\mathcal{J}$, the set of robots $\mathcal{I}$, and the reward for any task-robot assignment are finite, the sum of rewards for any valid allocation is finite. Thus, there exists a real number $S_{\max} < \infty$ such that $S(\mathcal{A}) \le S_{\max}$ for all feasible allocations $\mathcal{A}$.
\end{proof}

\begin{lemma}
\label{lemma:jmax}
Let $V_{\max}$ be the maximum intrinsic reward of any single task and $Q_{\max}$ be the maximum support reward for a single task. The maximum global objective value $S_{\max}$ scales linearly with the number of tasks $N_t$.
\end{lemma}

\begin{proof}
The global objective function is the sum of rewards for all $N_t$ tasks. Since only one robot can be assigned to a task, each task contributes at most one intrinsic term and one support term:
\begin{equation}
    S(\mathcal{A}) = \sum_{j=1}^{N_t} \left( R_{intrinsic}(j) + R_{support}(j) \right).
\end{equation}
Substituting the maximal bounds:
\begin{equation}
    S_{\max} \le \sum_{j=1}^{N_t} (V_{\max} + Q_{\max}) = N_t \cdot (V_{\max} + Q_{\max}).
\end{equation}
Since $V_{\max}$ and $Q_{\max}$ are constant system parameters independent of input size, it follows that $S_{\max}$ scales as $O(N_t)$.
\end{proof}

\begin{theorem}
\label{thm:convergence}
The iterative auction algorithm converges in a finite number of steps $K_{conv}$, which is bounded by $O(N_t / \epsilon)$.
\end{theorem}

\begin{proof}
By Property \ref{prop: mono}, the algorithm monotonically increases the objective function by at least $\epsilon$ at each step. Summing this improvement over $K$ iterations gives:
\begin{equation}
    S(\mathcal{A}_K) \ge S(\mathcal{A}_0) + K \cdot \epsilon.
\end{equation}
We assume allocations are initialized from scratch (i.e., $\mathcal{A}_0 = \emptyset$ with $S(\emptyset) = 0$). Since the objective value is bounded by $S_{\max}$ (Lemma \ref{lemma:boundedness}), we conclude:
\begin{equation}
    K_{conv} \le \frac{S_{\max}}{\epsilon}.
\end{equation}
Lemma \ref{lemma:jmax} then implies that $K_{conv}$ scales as $O(N_t / \epsilon)$.
\end{proof}
\subsection{Complexity Analysis}

\begin{lemma}
\label{lemma:per_iter}
For each iteration, determining the winning bid has a time complexity of
$O\!\left(N_t^2 N_r^2 N_s \left(N_h^2 + N_t\right)\right)$.
\end{lemma}

\begin{proof}
In each auction iteration, all $N_r$ robots evaluate bids for all $N_t$ tasks at up to $N_t$ possible insertion positions within their current task sequence. This results in $O(N_t^2 N_r)$ candidate configurations per iteration.

For each candidate configuration, the global objective must be recomputed. For all $N_s$ samples, we construct $\mathbf{r}$ by iterating over up to $N_h$ HVUT failures, finding the earliest robot and updating all future $\mathbf{t}$ for this sample. Additionally, up to $N_t$ tasks may need to be evaluated for all $N_r$ robots to compute the corresponding marginal gain. Combining these factors yields a total per-iteration complexity of $O\!\left(N_t^2 N_r^2 N_s \left(N_h^2 + N_t\right)\right)$.
\end{proof}

\begin{theorem}
The proposed algorithm converges in pseudo-polynomial time. Specifically, the total time complexity is $O\!\left(\frac{1}{\epsilon}N_t^3 N_r^2 N_s \left(N_h^2 + N_t\right)\right)$.
\end{theorem}

\begin{proof}
The total computational cost $T_{total}$ is the product of the maximum iterations (Theorem \ref{thm:convergence}) and the cost per iteration (Lemma \ref{lemma:per_iter}):
\begin{equation}
    T_{total} = O\!\left(\frac{1}{\epsilon}N_t^3 N_r^2 N_s \left(N_h^2 + N_t \right)\right)
\end{equation}
\end{proof}

\subsection{Comments on Solution Quality}
Although recent work has established bounds for non-submodular functions under cardinality constraints \cite{das2018approximate} or via randomized methods \cite{chen2018weakly}, deriving an approximation lower bound for general greedy methods over non-submodular functions with general matroid partition constraints (as in MRTA) remains an open theoretical challenge. As a result, we rely on empirical evaluation against several established baseline methods to validate solution quality.

\begin{figure*}[t]
    \centering
    \includegraphics[width=1.0\textwidth, keepaspectratio]{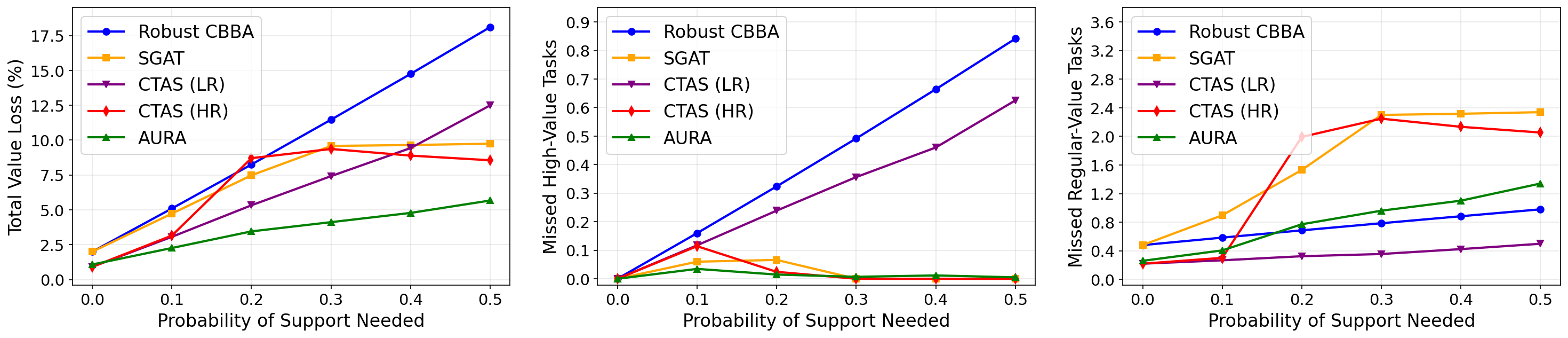}
    \vspace{-0.6cm}
    \caption{Total expected value loss and average missed tasks for an 8-robot, 12-task (4 HVUTs) mission with known uncertainty, where the uncertainty realization time is $\mathcal{T}=50$. We compare performance across a range of probabilities that search tasks require debris-clearing assistance. The results show that AURA consistently achieves the lowest value loss, meeting nearly all high-value deadlines while completing more regular-value tasks within their deadlines than redundancy-based approaches.}
    \label{fig:robust_by_prob}
    \vspace{-0.3cm}
\end{figure*}

\section{Simulation and Results}

\subsection{Simulation Setup}
To evaluate the performance of AURA, a simulated disaster relief scenario is modeled from Fig. \ref{fig:disaster_example}, where there exists a probability $\mathcal{P}$ that search tasks require the debris-clearing capability an estimated $\mathcal{T}$ seconds after the initial arrival. Each task has a deadline $\mathcal{D}$, after which no reward is obtained. Search tasks have value $\bar{v}=400$, while debris tasks have value $\bar{v}=100$. We use a time-discount factor of $\lambda=0.99$ to encourage earlier arrival times while still retaining nearly full value when deadlines are met. We use $N_s=50$ samples (enumerating all outcomes when $2^{|\mathcal{H}|} < N_s$) and set $\epsilon=1.0$ to prevent insignificant task swaps.

Robots are initialized at the center of a $2{,}000 \times 2{,}000$ m environment with tasks randomly distributed throughout the region. Robot speeds are fixed at 5 m/s, and all task durations are set to 300 s. To model sequential task completion, deadlines are generated in batches. For the first batch, deadlines depend on the distance from the start location, the task duration, and a random slack term. For subsequent batches, deadlines additionally account for the previous batch timing and a heuristic inter-task travel time estimate. For search tasks, slack is sampled from $[\mathcal{T}, \mathcal{T}+200]$ to create urgency while allowing time for potential support. Debris task slack is sampled from $[0,400]$. If support is required, search progress halts until a debris robot arrives and clears the obstruction, requiring an additional 50 s.

% (Maybe table for capabilities? Path-clearing have slower speed?). 

\subsection{Baselines}
We benchmark AURA against several baseline methods that differ in both algorithmic design and uncertainty handling. These methods include:
\begin{itemize}[leftmargin=*, itemsep=2pt, topsep=2pt]
    \item \textit{CTAS} \cite{fu_robust_2022}: solves a two-stage stochastic program to minimize task failure risk under task requirement uncertainty. For fair comparison, we adapt the original formulation by incorporating task deadline constraints, modeling uncertainty with Bernoulli distributions (instead of Gaussian), and penalizing individual task arrival times rather than the total mission time. 
    Additionally, a near-zero conditional value at risk (CVaR) coefficient ($\alpha = 0.001$) is used to approximate expected-value optimization. We evaluate two risk-penalty settings: a high risk penalty (HR), which induces greater redundancy, and a low risk penalty (LR), which more evenly balances risk and deadline satisfaction.
    \item \textit{Robust CBBA} \cite{Ponda2012Thesis}: approximates expected reward via sampling of uncertain task outcomes. Although originally developed for decentralized settings, we use this method as a state-of-the-art sampling-based greedy baseline and implement it on a centralized node for fair comparison.
    \item \textit{Sequential Greedy + Auxiliary Tasks (SGAT)}: a modified sequential greedy algorithm that augments the task set with the auxiliary tasks defined in Section \ref{subsec:aux}. By treating support actions as real tasks with utilities equal to their expected support value, this baseline serves as a conservative greedy approach that allows redundant assignments.
\end{itemize}

\subsection{Simulation Results}

\textbf{Robust Experiments}. In these experiments, the true capability distribution for search tasks is revealed $\mathcal{T}=50$ s after arrival, simulating time for the robots to confirm whether the additional debris-clearing capability is required. Fig. \ref{fig:robust_by_prob} illustrates the expected performance for a mission with 8 robots (4 search, 4 debris) and 12 tasks (4 search, 8 debris) across 50 random environments. The left plot shows the total percent value loss relative to meeting all deadlines while the middle and right plots report the expected number of missed deadlines for high-value search tasks and regular-value debris tasks, respectively.

We observe that the less proactive methods, Robust CBBA and CTAS (LR), meet the most regular-value deadlines but miss the largest number of high-value deadlines. Because these approaches do not anticipate capability uncertainty, support robots are not positioned to respond in time if the debris-clearing capability is required, leading to high-value deadline violations. Although CTAS (LR) achieves good performance, this is largely attributable to the optimality of its underlying solver rather than effective handling of uncertainty, as evident in the middle plot of Fig. \ref{fig:robust_by_prob}.

The redundancy-based methods significantly reduce the number of missed high-value deadlines, completely eliminating them for all but low-uncertainty scenarios. However, this comes at the expense of additional missed regular-task deadlines. This is because over-committing support robots consumes resources that could otherwise complete debris tasks in many cases where the debris-clearing capability is not required.

In contrast, AURA achieves the strongest overall performance. Across all probabilities, we meet nearly all high-value deadlines while remaining competitive with Robust CBBA in terms of regular-task completion. This combined performance is reflected in the first plot of Fig. \ref{fig:robust_by_prob}, where AURA \textit{consistently incurs the lowest total value loss among all approaches.} This improvement stems from strategically assigning support robots to tasks near HVUTs, balancing task progress with quick intervention when support is required.

\textbf{Resilient Experiments.} We evaluate two values of $\beta$ from Eq. (4), which controls how quickly task feasibility decays. A large value ($\beta=0.01$) causes feasibility to decrease fast, shortening the window in which the search robot can attempt recovery (e.g., exploring alternative routes) and leaving less time for a support robot to proactively adjust its position. A smaller value ($\beta=0.004$) results in slower decay, leaving more time for exploring alternative routes and for proactive positioning of a support robot. Once task feasibility begins to decrease, we replan at a fixed frequency, evaluating the tradeoff between maintaining the current assignment and repositioning for closer support. 

To model a variety of scenarios with unexpected debris, we introduce a sampled parameter $f_\text{floor}\in[0.3,0.8]$, which represents the feasibility level at which the robot recovers when failure does not occur. For example, when a search robot unexpectedly encounters debris, explores alternative routes, and ultimately finds another route to complete the task, task feasibility decreases until reaching $f_\text{floor}$ and then returns to one, indicating that the search task can continue without a debris robot. On the other hand, when feasibility reaches zero, this implies that the robot failed to find another route in the designated recovery time and that a debris robot is required. To approximate performance in expectation, we sample successes and failures assuming that $f_\text{floor}$ represents the underlying probability of success. For example, if $f_\text{floor}=0.4$, we expect 40\% of samples to be successes (no support needed) and 60\% of samples to require support.

Table \ref{tab:impact_decay_single} reports the average value loss and missed tasks for 40 randomly sampled environments with 6 robots (3 search, 3 debris) and 10 tasks (3 search, 7 debris). We observe that proactively considering reallocations when search robots are given the opportunity to recover from unexpected debris obstruction greatly improves overall mission performance. Similar to the robust experiments, the redundancy-based methods satisfy a greater number of high-value deadlines, whereas the less proactive methods complete more regular-value tasks at the cost of high-value deadline violations. Across both slow and fast decay scenarios, AURA \textit{achieves the lowest value loss}, demonstrating the benefit of proactively repositioning a support robot to nearby debris tasks when possible. As expected, the larger improvement occurs when search robots are given more time to attempt recovery as it provides more time for proactive repositioning.

\begin{table}[t]
\centering
\caption{Performance of a 6-robot, 10-task (3 HVUTs) mission with unexpected uncertainty under varying decay settings}
\label{tab:impact_decay_single}
\renewcommand{\arraystretch}{1.2}
\setlength{\tabcolsep}{5.5pt} % Adjust spacing if needed
\footnotesize % Ensures correct IEEE table body font size

\begin{tabular}{lcccc}
\toprule
\multirow{2}{*}{\textbf{Algorithm}} 
 & \multicolumn{2}{c}{\textbf{Value Loss (\%)}} 
 & \multicolumn{2}{c}{\textbf{Missed Tasks (HV / Reg.)}} \\ 
\cmidrule(r){2-3} \cmidrule(l){4-5} % Cleanly divides the subheadings
 & \textbf{Fast} & \textbf{Slow} 
 & \textbf{Fast} & \textbf{Slow} \\ 
\midrule
 
Robust CBBA & 20.95 & 23.12 & 0.83 / 0.66 & 0.95 / 0.59 \\

SGAT & 10.79 & 10.52 & 0.14 / 1.49 & 0.11 / 1.56 \\

CTAS (LR) & 16.38 & 17.32 & 0.65 / \textbf{0.46} & 0.72 / \textbf{0.41} \\

CTAS (HR) & 7.26 & 7.42 & \textbf{0.00} / 1.38 & \textbf{0.00} / 1.41 \\ 

AURA & \textbf{6.58} & \textbf{5.31} & \textbf{0.00} / 1.25 & \textbf{0.00} / 1.01 \\ 
\bottomrule
\end{tabular}
\vspace{-0.2cm}
\end{table}

\begin{figure*}[t]
    \centering
    \includegraphics[width=0.99\textwidth, keepaspectratio]{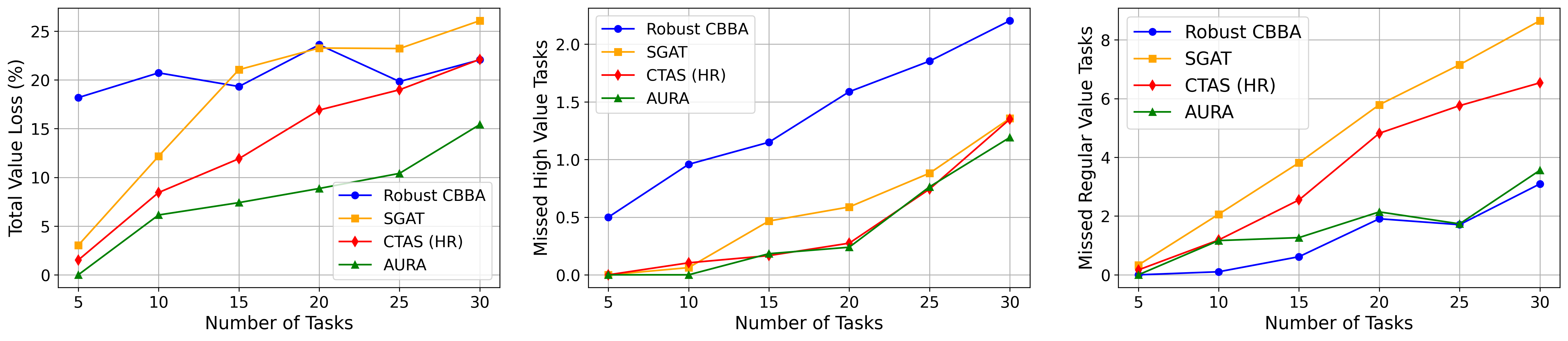}
    \vspace{-0.2cm}
    \caption{Total value loss and number of missed tasks as the problem size increases. We assume a fixed team of 3 search robots and 3 debris robots and maintain a ratio of approximately 2:1 debris tasks to search tasks. As the number of tasks grows, AURA consistently achieves the best performance, satisfying as many HVUT deadlines as the redundancy-based methods while maintaining a level of regular-value deadline completion comparable to the non-proactive approach.}
    \label{fig:scalability}
    \vspace{-0.3cm}
\end{figure*}

\begin{figure}[t]
    \centering
    \includegraphics[width=.35\textwidth, keepaspectratio]{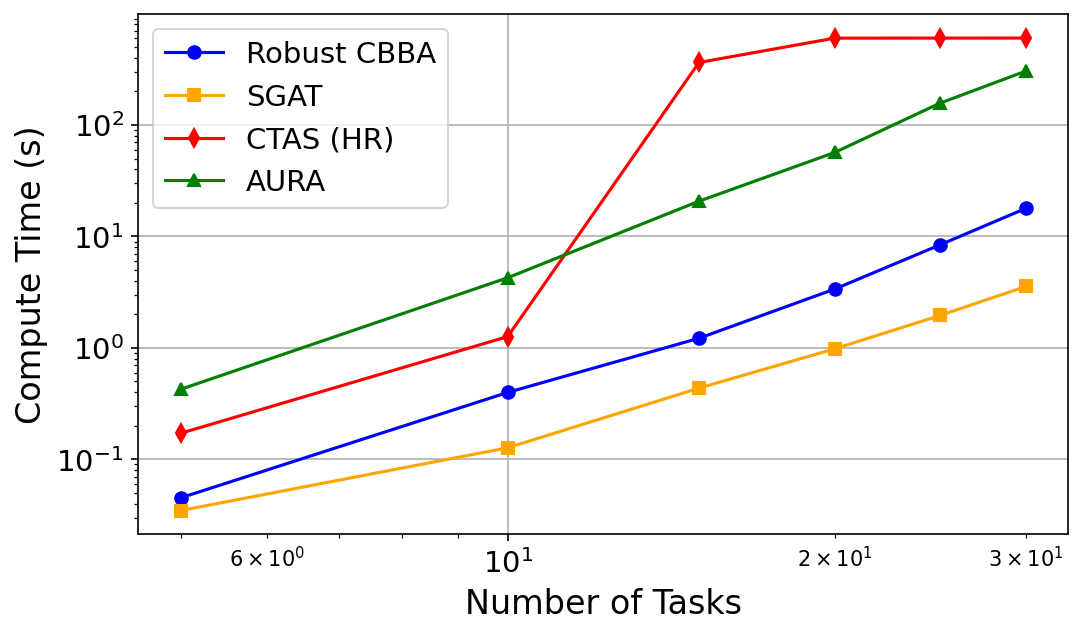}
    \vspace{-0.3cm}
    \caption{Runtime analysis as the number of tasks increases. On a log-scale plot, all greedy methods exhibit linear growth, indicating the expected polynomial-time complexity. In contrast, CTAS exhibits super-linear growth and reaches the 600-second time limit for larger problem sizes. }
    \label{fig:time_complexity}
    \vspace{-0.3cm}
\end{figure}

\textbf{Scalability.} Fig. \ref{fig:scalability} illustrates the performance scalability of AURA across increasing problem sizes. We fix a team of 6 robots (3 search, 3 debris) and maintain an approximate 2:1 ratio of debris tasks to 
search tasks as the total number of tasks increases. For each problem size, we generate 10 random environments and randomly sample the probability of the debris-clearing capability being required for search tasks from the interval $[0.2, 0.5]$. Because computing exact expected values becomes intractable for larger missions, we sample 20 realizations from the joint uncertainty distribution and average the resulting outcomes. Furthermore, since CTAS requires solving a mixed-integer linear program (MILP), we impose a 600-second time limit on its execution and elect to evaluate only the CTAS (HR) variant as it exhibits the proactive behavior of interest. 

Despite the need to model sequential high-value task completion in these experiments, AURA still consistently achieves the lowest total value loss, while the baseline methods exhibit trends similar to those observed in earlier experiments. One notable change is that SGAT performs the worst in larger instances, which can be attributed to its non-coupled formulation. In some cases, SGAT assigns debris robots to search tasks that are not scheduled to be executed for some time, ultimately abandoning them after waiting for a fixed duration.

Fig. \ref{fig:time_complexity} shows a runtime analysis across the same experiments as Fig. \ref{fig:scalability}. All experiments were conducted on a machine equipped with an Intel i9-10920X CPU, and Gurobi is used as the solver for CTAS. On this logarithmic plot, AURA exhibits linear growth, suggesting that its empirical performance is consistent with the derived pseudo-polynomial bound. The two greedy baselines achieve lower computation times due to their reduced objective complexity. In contrast, CTAS exhibits super-linear growth and reaches the 600-second time limit for instances with 20 tasks or more, indicating its exponential complexity. 

\section{Conclusion}
 This paper presents an auction-based algorithm with pseudo-polynomial convergence guarantees for MRTA under task requirement uncertainty, along with a novel unified framework capable of handling both modeled and unmodeled uncertainty. Our results demonstrate that mission value can be preserved by valuing tasks not only for their immediate reward, but also for the future reward created by intelligent positioning near uncertain tasks. Furthermore, by allowing for proactive reallocations when unexpected uncertainties arise, AURA yields consistent gains over purely reactive strategies. Finally, AURA is scalable, demonstrating continued performance advantages in larger missions. 

 \section{Acknowledgments}
The authors thank Siddharth Mayya for many valuable discussions throughout the development process. The authors also acknowledge Nano Banana Pro for assistance in generating the icons used in Fig. \ref{fig:disaster_example}.

\bibliographystyle{IEEEtran}
\bibliography{references}
\addtolength{\textheight}{-12cm}   % This command serves to balance the column lengths

%%%%%%%%%%%%%%%%%%%%%%%%%%%%%%%%%%%%%%%%%%%%%%%%%%%%%%%%%%%%%%%%%%%%%%%%%%%%%%%%

%%%%%%%%%%%%%%%%%%%%%%%%%%%%%%%%%%%%%%%%%%%%%%%%%%%%%%%%%%%%%%%%%%%%%%%%%%%%%%%%

%%%%%%%%%%%%%%%%%%%%-%%%%%%%%%%%%%%%%%%%%%%%%%%%%%%%%%%%%%%%%%%%%%%%%%%%%%%%%%%%%

%%%%%%%%%%%%%%%%%%%%%%%%%%%%%%%%%%%%%%%%%%%%%%%%%%%%%%%%%%%%%%%%%%%%%%%%%%%%%%%%

\end{document}